\documentclass[runningheads]{llncs}

\usepackage[mobile]{eccv}

\usepackage{eccvabbrv}

\usepackage{graphicx}
\usepackage{booktabs}
\usepackage{soul}
\usepackage{colortbl}

\usepackage{capt-of}   
\usepackage{xcolor}
\definecolor{TeaserBlue}{RGB}{120,170,255}
\definecolor{TeaserGreen}{RGB}{80,200,120}
\definecolor{TeaserOrange}{RGB}{255,165,70}

\usepackage[accsupp]{axessibility}  

\usepackage{hyperref}

\usepackage{orcidlink}

\begin{document}

\title{ONE-SHOT: Compositional Human-Environment Video Synthesis via Spatial-Decoupled Motion Injection and Hybrid Context Integration} 

\titlerunning{ONE-SHOT}
\begingroup
\makeatletter
\def\@thefnmark{$*$}
\@footnotetext{Corresponding author.}
\def\@thefnmark{$^\dagger$}
\@footnotetext{Project leader.}
\makeatother
\endgroup

\author{Fengyuan Yang\inst{1,2} \and
Luying Huang\inst{2}$^\dagger$ \and
Jiazhi Guan\inst{2}$^*$ \and
Quanwei Yang\inst{2} \and
Dongwei Pan\inst{2} \and
Jianglin Fu\inst{2} \and
Haocheng Feng\inst{2} \and
Wei He\inst{2} \and
Kaisiyuan Wang\inst{2} \and
Hang Zhou\inst{2}$^*$ \and
Angela Yao\inst{1}}

\authorrunning{F.Yang et al.}

\institute{
National University of Singapore
\and
Baidu Inc.
}

\maketitle

\begin{figure}[h!]
\includegraphics[width=1.0\linewidth]{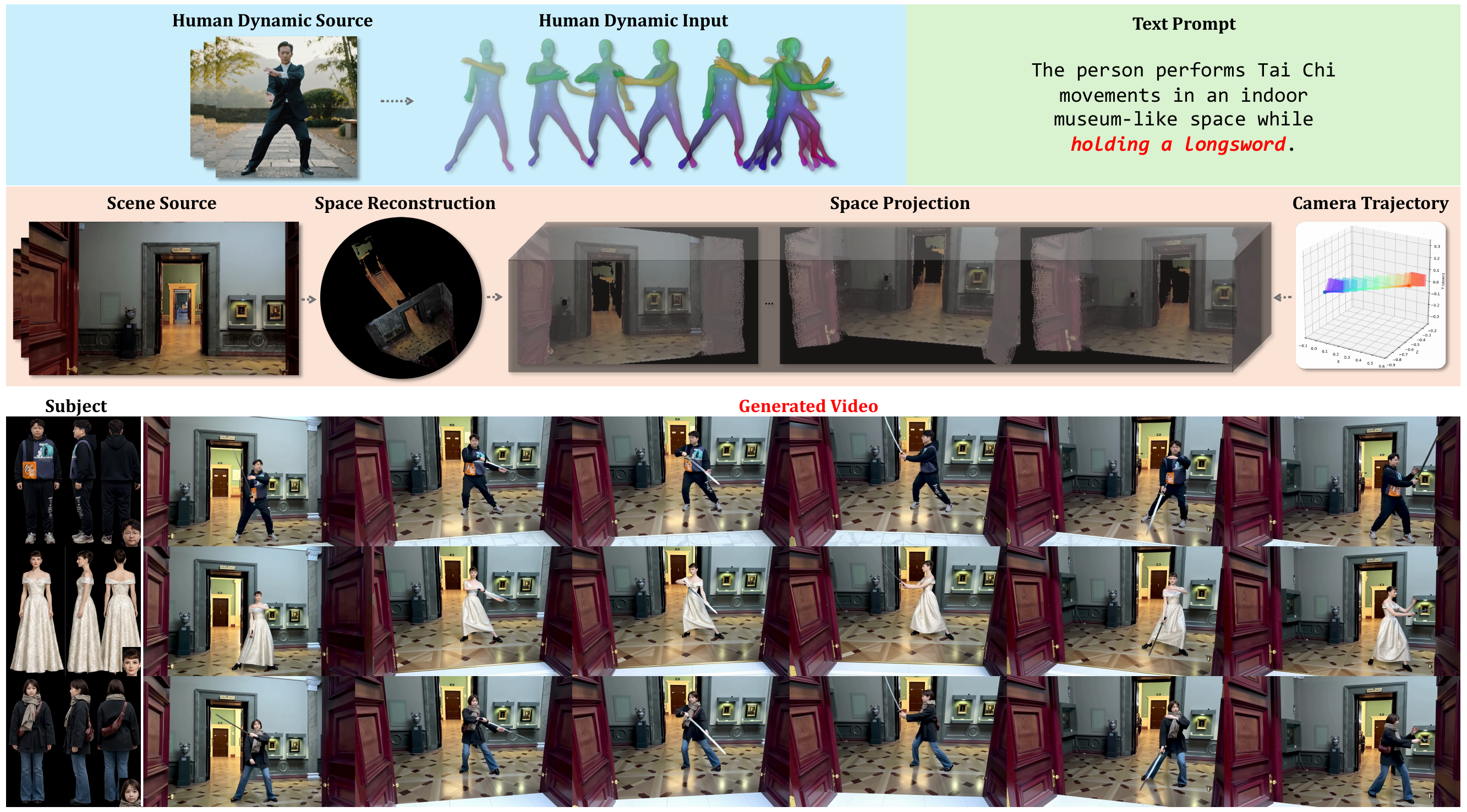}
\caption{
\textbf{Human-Environment Synthesis by ONE-SHOT}.
Three kinds of conditions are decoupled and factorized within our ONE-SHOT framework to enable compositional generation.
\textcolor{TeaserBlue}{\textit{Blue})}: human dynamics. 
\textcolor{TeaserGreen}{\textit{Green})}: textual prompt. 
\textcolor{TeaserOrange}{\textit{Orange})}: environment space.
The generated videos exhibit high consistency in human subjects and environments \textit{w.r.t.} given conditions, and further accurately convey the textual interaction of ``holding a longsword''.
}
 \label{fig:teaser}
\end{figure}

\begin{abstract}


Recent advances in Video Foundation Models (VFMs) have revolutionized human-centric video synthesis, yet fine-grained and independent editing of subjects and scenes remains a critical challenge.
Recent attempts to incorporate richer environment control through rigid 3D geometric compositions often encounter a stark trade-off between precise control and generative flexibility. 
Furthermore, the heavy 3D pre-processing still limits practical scalability.
In this paper, we propose ONE-SHOT, a parameter-efficient framework for compositional human-environment video generation. 
Our key insight is to factorize the generative process into disentangled signals. 
Specifically, we introduce a canonical-space injection mechanism that decouples human dynamics from environmental cues via cross-attention. 
We also propose Dynamic-Grounded-RoPE, a novel positional embedding strategy that establishes spatial correspondences between disparate spatial domains without any heuristic 3D alignments. 
To support long-horizon synthesis, we introduce a Hybrid Context Integration mechanism to maintain subject and scene consistency across minute-level generations. 
Experiments demonstrate that our method significantly outperforms state-of-the-art methods, offering superior {structural control and creative diversity} for video synthesis.
Our project has been available on: \url{https://martayang.github.io/ONE-SHOT/}.

\keywords{Controllable video generation \and human-centric video synthesis \and diffusion models \and video diffusion}
\end{abstract}
\section{Introduction}
\label{sec:intro}

The surge in visual content creation in recent years has been catalyzed in part by Video Foundation Models (VFMs) such as Wan~\cite{wanvideo} and HunyuanVideo~\cite{hunyuanvideo}.
These models empower sophisticated capabilities in animation, video editing, and generative synthesis~\cite{Dreamactorm1,vace,wei2025univideo}. 
Particularly in human-centric applications~\cite{opens2v,phantom,hu2023animate,echomimicv2,xu2024hallo,cheng2025wan}, these models are increasingly indispensable across diverse fields such as e-commerce, digital journalism, education, and multimedia entertainment.

Certain efforts~\cite{vace,phantom,li2025bindweave,wei2025univideo,cheng2025wan} have sought to extend these capabilities to include subject and environment editing through million-scale supervised fine-tuning. However, these approaches remain limited in their precision. They often lack the capacity to separate and independently control human movement and the environmental background, limiting their utility in professional workflows that demand highly customized and structurally consistent content. We argue that leveraging 3D signals to precisely regulate both human and scene information is of paramount importance, as they provide the necessary geometric grounding to achieve decoupled control and ensure high-fidelity spatial interactions.



%

Recently, a few studies have also identified that 
the use of explicit 3D geometry information allows for more fine-grained motion control and 
expands generative freedom by modeling interactions between subjects and the environment. 
For instance, Uni3C~\cite{cao2025uni3c} introduces a unified framework that fuses human motion signals with point clouds, enabling high-fidelity video generation with controllable camera trajectories. Similarly, RealisMotion~\cite{liang2025realismotion} decouples character motion from {environment dynamics} at the input level, facilitating independent manipulation of both the subject's actions and the background scene.

Despite these advancements, {several critical bottlenecks hinder broader application}:
\textbf{1)} Burdensome 3D Processing: Current methods rely heavily on the explicit alignment of complex spatial signals (\textit{e.g.}, point clouds) with human motion. This requirement for unified 3D pre-processing imposes significant constraints on scalability and real-world applicability.
\textbf{2)} Compromised Generative Abilities: Intensive fine-tuning on narrow downstream tasks with rigid conditioning often compromises the inherent expressive power of VFMs. This results in a degradation of creative diversity.
\textbf{3)} Short-term Synthesis: Existing frameworks are largely restricted to short-duration synthesis ($\sim$10s), failing to maintain global spatio-temporal consistency over long-horizon sequences within a persistent 3D environment and subject appearance.

To address these challenges, 
we propose a c\textbf{O}mpositional huma\textbf{N}-\textbf{E}nvironment video synthesis framework based on \textbf{S}patial-decoupled motion injection and \textbf{H}ybrid c\textbf{O}ntext in\textbf{T}egration, namely \textbf{ONE-SHOT}.
\textit{
Our key insight lies in factorizing the {generation process into a compositional synthesis} of distinct signals—human motion, environmental geometry, and camera trajectories—integrated within a unified, coherent framework.
}
A key technical bottleneck in existing methods is the {implicit spatial coupling} of motion and scene signals, which often leads to the ``\textit{over-conditioning}'' phenomenon in which denser structural input dominates the generative process. 
{By decoupling} human motion priors from environmental geometry via a canonical-space injection mechanism, we effectively mitigate this conditioning competition,
thereby preserving the original semantic flexibility within pretrained VFMs.
As illustrated in Fig.~\ref{fig:teaser}, this compositional strategy enables the {synthesis of vivid videos with persistent appearance}, offering a practical balance between {precise control and generative flexibility}.

Specifically, our architecture adopts a ControlNet-style design~\cite{zhang2023adding} to leverage the extensive pre-trained knowledge of VFMs. 
%
Unlike previous methods~\cite{cao2025uni3c,liang2025realismotion} that spatially fuse human motion with environmental cues, {which can impose excessively restrictive constraints during training}, 
{our approach injects human motions through a dedicated cross-attention layers in a canonical space}. 
This decoupled injection mechanism offers a sparser condition tuning, mitigating the risk of ``collapse'' of the VFM's inherent expressiveness. 
{Furthermore}, rather than updating the entire conditioning branch, we employ a parameter-efficient optimization strategy using LoRA~\cite{hu2022lora}. 
We observe that this minimalist tuning strategy is sufficient for precise control.

One key advantage of the stand-alone human motion injection mechanism is that the need for {complex, heuristic} 3D pre-processing is eliminated.
Our pipeline first synthesizes an environment-only sequence via 2D point cloud projections under a controllable camera trajectory. 
We then overlay a human layout onto this sequence to define the subject's coarse spatial position. 
{This design bypasses the complexity of coordinate alignment between disparate spatial domains of the environment and the human body, significantly enhancing generative flexibility.}
To address the spatial misalignment between the two domains, we propose 
\textit{\textbf{Dynamic-Grounded-RoPE}} 
in cross-attention. This mechanism adaptively scales rotary positional embeddings, establishing accurate spatial correspondences between the environment and the canonical human pose input.

Moreover, to maintain spatio-temporal consistency for minute-level generation, our framework leverages a 
\textit{\textbf{Hybrid Context Integration}}
mechanism: 
static context reference tokens to preserve subject identity and dynamic context memory tokens to track the overall appearance. 
These features collectively provide sufficient context to prevent identity drift and environmental artifacts, enabling long-shot consistent human–environment video generation over minutes.

Our contributions are summarized as follows:
\textbf{1)} We introduce ONE-SHOT, a parameter-efficient framework built upon pre-trained VFMs. By optimizing only a sparse set of parameters, it achieves high-fidelity synthesis of human-environment videos with independent control over subject appearance, human dynamics, spatial environments, and camera trajectories.
\textbf{2)} We propose a canonical-space motion injection mechanism that better mitigates the conditioning competition between rigid human priors and text prompting. This mechanism ensures precise control and also preserves responsiveness to textual instructions.
\textbf{3)} By anchoring static and dynamic context, our method ensures persistent subject identity and stable human-environment interactions across minute-scale generations.
\textbf{4)} Extensive experiments and ablation studies demonstrate that our approach significantly outperforms existing methods, offering superior structural control and better creative diversity.
\section{Related Work}
\label{sec:related_work}



The rapid evolution of large-scale video generation models~\cite{hunyuanvideo, wanvideo, hacohen2024ltx, CogVideoX, opensora} has catalyzed the development of general-purpose editing frameworks~\cite{vace, wei2025univideo, burgert2025motionv2veditingmotionvideo, lei2025ditraj, kulikov2025flowedit}.
Despite their versatility, these methods often treat human subjects as holistic pixel regions, lacking the sub-pixel physical accuracy and structural awareness required for complex skeletal deformations. 
This limitation motivates a shift toward more specialized approaches that incorporate explicit human structural priors and 3D environmental constraints.

\subsection{Human-Centric Reenactment}

To achieve more granular authority over human forms, human-centric reenactment~\cite{Wan-Animate, RealisDance-DiT, echomimicv2, meng2025echomimicv3, Champ, Dreamactorm1, zhang2025mimicmotion, hu2023animate, hu2025animateanyone2} leverages structured motion priors to drive realistic character animations. 
Early efforts like Champ~\cite{Champ} incorporated explicit SMPL-based 3D shape priors to improve robustness against large viewpoint changes, while Animate Anyone 2~\cite{hu2025animateanyone2} set a new benchmark for identity preservation through large-scale video pre-training. 
More recently, building upon the Wan foundational model~\cite{wanvideo}, Wan-Animate~\cite{Wan-Animate} employs spatially-aligned skeleton signals and implicit facial features within a multi-stage training pipeline to achieve superior controllability and expressiveness. 
Simultaneously, Realisdance-DiT~\cite{RealisDance-DiT} emphasizes motion reliability by introducing targeted modifications to the DiT architecture and employing flexible fine-tuning strategies, which significantly enhances stability in rare poses and complex character-object interactions. 
While these methods achieve remarkable 2D fidelity, they primarily animate characters in isolated or static settings, failing to perceive or respond to the physical constraints of complex 3D environments.

\subsection{Human-Environment Video Synthesis}

Moving beyond character animation in static settings, human-en\-vironmental video synthesis~\cite{cao2025uni3c, liang2025realismotion} addresses the challenge of grounding movements within explicit 3D scenes to model complex interactions.
Uni3C~\cite{cao2025uni3c} achieves unified 3D coordination by explicitly aligning human motion with scene geometry(e.g., point clouds) to resolve artifacts like interpenetration. 
RealisMotion~\cite{liang2025realismotion} further introduces physical priors within a decoupled world-space control framework, significantly mitigating foot-skating and enhancing the dynamic realism of interactions. 
To further unlock generative freedom, works like MotionCtrl~\cite{wang2024motionctrl} and 3DTrajMaster~\cite{fu20243dtrajmaster} incorporate controllable camera trajectories and 3D viewpoints, enabling the synthesis of complex interactions from dynamic perspectives within a persistent 3D space. 
Despite these advancements, current frameworks remain hindered by burdensome 3D pre-processing, a loss of generative diversity due to rigid conditioning, and a failure to maintain spatio-temporal consistency over long-horizon sequences.
To address these challenges, our proposed {ONE-SHOT} disentangles human dynamics from spatial environments, which simultaneously preserves generative diversity and persistent subject consistency.
\section{Preliminaries}
\label{sec:preliminaries}

\noindent \textbf{Video Foundation Models}.
We build our model upon a pretrained video foundation model (VFM) \textbf{Wan2.1}~\cite{wanvideo} that is trained with flow matching in a latent space.
Given an input video, a VAE encoder maps it into a latent tensor $\mathbf{z}_0\in\mathbb{R}^{T\times H\times W\times C}$, where $T$ is the number of frames, $H\times W$ denotes the spatial token grid, and $C$ is the channel dimension. The VFM learns a time-dependent velocity field that transports a noise latent to the data latent.
Concretely,  a noise latent $\mathbf{z}_1\sim\mathcal{N}(\mathbf{0},\mathbf{I})$ and a timestep $t\in[0,1]$ sampled from a logit-normal distribution are used to construct an intermediate latent $\mathbf{z}_t = t\,\mathbf{z}_1 + (1-t)\,\mathbf{z}_0$.
Following Rectified Flows~\cite{esser2024scalingrectifiedflowtransformers}, the ground-truth velocity along this path is
\begin{equation}
\mathbf{v}_t = \frac{d\mathbf{z}_t}{dt} = \mathbf{z}_1 - \mathbf{z}_0 .
\label{eq:rf_velocity_z}
\end{equation}
A conditional velocity predictor $\mathbf{v}_\theta$ is trained by minimizing the mean squared error (MSE)
\begin{equation}
\mathcal{L}_{\mathrm{FM}}
=
\mathbb{E}_{\mathbf{z}_0,\mathbf{z}_1,t}
\left[
\left\|
\mathbf{v}_\theta(\mathbf{z}_t,t,\mathbf{c}) - \mathbf{v}_t
\right\|_2^2
\right],
\label{eq:fm_loss_z}
\end{equation}
where $\mathbf{c}$ denotes conditioning signals. 


\noindent \textbf{Rotary Position Embedding}.
Rotary Position Embedding (RoPE) encodes token positions by applying a position-dependent rotation to paired query/key channels before attention.
For a token at position $\mathbf{p}$ (e.g., $\mathbf{p}=(t,x,y)$ on the latent grid), RoPE can be written abstractly as
\begin{equation}
\tilde{\mathbf{Q}} = \mathcal{R}(\mathbf{p})\,\mathbf{Q},\qquad
\tilde{\mathbf{K}} = \mathcal{R}(\mathbf{p})\,\mathbf{K},
\label{eq:rope}
\end{equation}
where $\mathcal{R}(\mathbf{p})$ is constructed from sinusoidal functions at multiple frequencies.
The rotated $\tilde{\mathbf{Q}},\tilde{\mathbf{K}}$ are then used in attention to make interactions sensitive to relative offsets between tokens.
In Sec.~\ref{sec:method}, we introduce a scale-grounded variant of RoPE tailored for our decoupled conditioning setup.

\begin{figure*}[!t]
\centering
\includegraphics[width=1.0\linewidth]{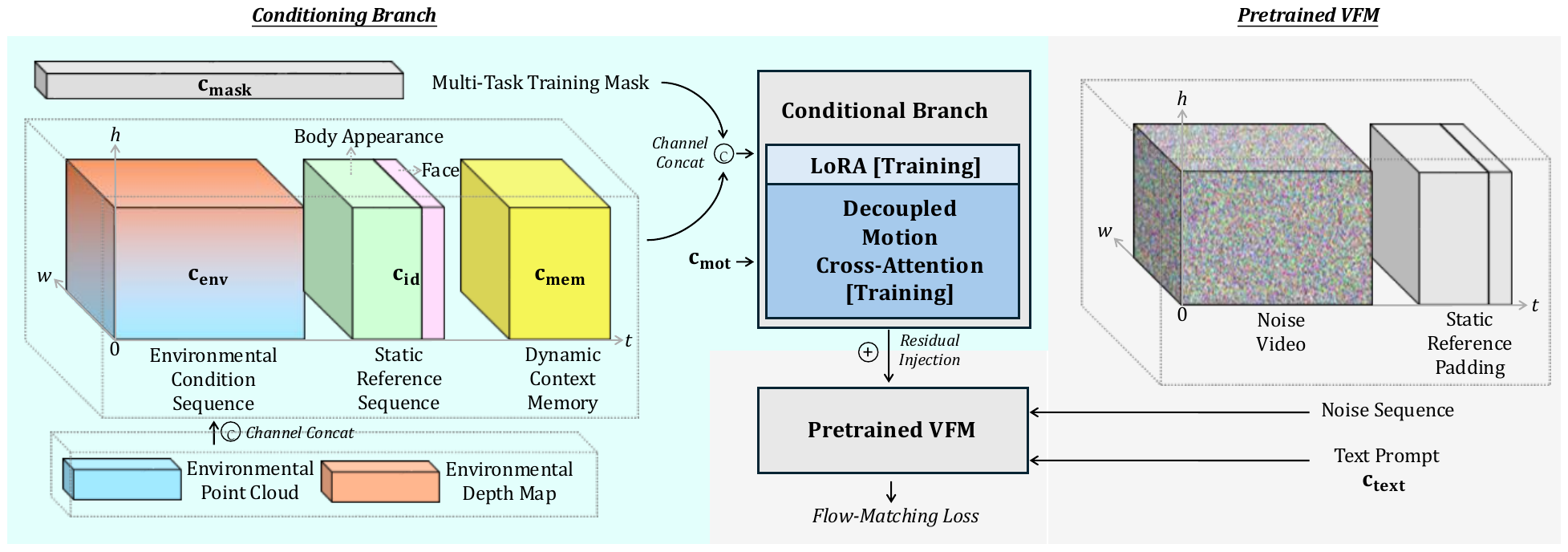}
\caption{
\textbf{Model Architecture.}
Our model is built upon a \textit{Pretrained VFM} (gray, right) augmented with an additional \textit{Conditioning Branch} (cyan, left) for concept injection.
The environmental condition $\mathbf{c}_{\text{env}}$ is encoded from 2D-projected point clouds and depth maps, while identity appearance $\mathbf{c}_{\text{id}}$ and context memory $\mathbf{c}_{\text{mem}}$ jointly maintain visual coherence.
Human dynamics $\mathbf{c}_{\text{mot}}$ are disentangled from environmental inputs and injected through the proposed Decoupled Motion Cross-Attention.
}
\label{fig:pipe}
\end{figure*}

\section{Methodology}
\label{sec:method}

\textbf{Task Definition.} 
Given a scene point cloud $\mathcal{P}$ (e.g., reconstructed from a source video or multi-view images), camera extrinsics $\{\mathbf{T}_t\}_{t=1}^{T}$, person identity appearance $\mathbf{y}^{\text{id}}$, optional scene context memory $\mathbf{y}^{\text{mem}}$, and a specified human motion sequence $\mathbf{m}_{1:T}$, our goal is to synthesize a realistic human--environment video where a reference identity performs a specified motion while remaining coherent with a 3D scene under a controllable camera trajectory.
The output is a video $\mathbf{x}_{1:T}$ (or its latent representation $\mathbf{z}_0\in\mathbb{R}^{T\times H\times W\times C}$):
\begin{equation}
\mathbf{x}_{1:T} \;=\; \mathcal{G}\!\left(\mathcal{P}, \{\mathbf{T}_t\}_{t=1}^{T}, \mathbf{c}_{\text{id}}, \mathbf{c}_{\text{mem}}, \mathbf{m}_{1:T}, \mathbf{c}_{\text{text}}\right),
\label{eq:task_def}
\end{equation}
where $\mathbf{c}_{\text{text}}$ denotes text-prompt tokens for instruction-based control, and 
{$\mathbf{c}_{\text{id}}$ embeds subject appearance $\mathbf{y}^{\text{id}}$ while $\mathbf{c}_{\text{mem}}$ embeds scene-context memory $\mathbf{y}^{\text{mem}}$; together, they maintain visual coherence.

\noindent \textbf{Decoupled Conditioning Framework.}
We factorize the conditioning set into six complementary components:
\begin{equation}
\mathbf{c} \;=\; \{\mathbf{c}_{\text{text}}, \mathbf{c}_{\text{env}}, \mathbf{c}_{\text{mot}}, \mathbf{c}_{\text{id}}, \mathbf{c}_{\text{mem}}, \mathbf{c}_{\text{mask}}\}.
\label{eq:cond_set}
\end{equation}
Specifically, (a) \emph{environmental conditions} $\mathbf{c}_{\text{env}}$ consist of per-frame RGB/depth projections of the 3D scene $\mathcal{P}$; (b) \emph{motion conditions} $\mathbf{c}_{\text{mot}}$ specify human dynamics via canonical-space tokens and a coarse layout bounding-box; (c) \emph{appearance conditions} $\mathbf{c}_{\text{id}}$ and (d) \emph{context memory} $\mathbf{c}_{\text{mem}}$ jointly preserve visual coherence across short and long-horizon sequences; and (e) a spatio-temporal \emph{binary mask} $\mathbf{c}_{\text{mask}}$ guides localized synthesis, as shown in Fig.~\ref{fig:pipe}. We leverage a lightweight {ControlNet-style} parallel branch to perform interactions between all the conditions and the frozen VFM backbone. Concretely, dense spatial/visual priors, including $\mathbf{c}_{\text{env}}$, $\mathbf{c}_{\text{id}}$, $\mathbf{c}_{\text{mem}}$, and $\mathbf{c}_{\text{mask}}$ are directly injected.
While a dedicated cross-attention pathway is designed for $\mathbf{c}_{\text{mot}}$ to achieve decoupling control of human dynamics.
All lightweight modules are optimized using a multi-task training strategy.

\subsection{Conditioning Branch}
\label{sec:cond_branch}

The left part of Fig.~\ref{fig:pipe} depicts a ControlNet-style side branch that conditions the frozen VFM by injecting control signals as hierarchical residuals additions.
We organize control cues into a unified conditioning context sequence and encode it into multi-scale residual features that are added to selected layers of the VFM.  

\noindent \textbf{Environmental Condition Sequence.}
We represent the scene and camera trajectory using 2D projections rendered from the point cloud $\mathcal{P}$ under the target extrinsics $\{\mathbf{T}_t\}_{t=1}^{T}$.
Let $\pi(\cdot;\mathbf{T}_t)$ denote a rendering/projection operator.
We define the environmental condition as an RGB--D sequence
\begin{equation}
\mathbf{c}_{\text{env}}
=
\big\{\mathbf{RGBD}^{\text{env}}_t \;=\; \pi(\mathcal{P}; \mathbf{T}_t)\big\}_{t=1}^{T},
\label{eq:env_proj}
\end{equation}

\noindent \noindent \textbf{Hybrid Context Integration.}
We construct (i) an identity condition $\mathbf{c}_{\text{id}}$ from a static reference segment
$\mathbf{y}^{\text{id}}_{1:T_{\text{id}}}$ (body and face), and (ii) an optional context memory
$\mathbf{c}_{\text{mem}}$ from a dynamically updated segment $\mathbf{y}^{\text{mem}}_{1:T_{\text{mem}}}$,
enabled only for long-horizon generation:
\begin{equation}
\mathbf{c}_{\text{id}} = E\!\left(\mathbf{y}^{\text{id}}_{1:T_{\text{id}}}\right),\qquad
\mathbf{c}_{\text{mem}} = E\!\left(\mathbf{y}^{\text{mem}}_{1:T_{\text{mem}}}\right),
\label{eq:app_def}
\end{equation}
where $E(\cdot)$ encodes input frames into conditioning tokens.
Notably, $\mathbf{y}^{\text{id}}_{1:T_{\text{id}}}$ contains appearance-only cues, consistent with our decoupled formulation.
For long sequences, we follow~\cite{zhao2025spatiavideogenerationupdatable} to retrieve
$\mathbf{y}^{\text{mem}}_{1:T_{\text{mem}}}$ from previously generated content under similar viewpoints,
which helps stabilize long-range appearance.

\noindent \textbf{Conditioning Context Sequence.}
Following standard conditioning designs~\cite{vace}, we concatenate environmental and appearance segments along the temporal axis and append a multi-task mask sequence $\mathbf{c}_{\text{mask}}$ (introduced in Sec.~\ref{sec:training}), then concatenate them channel-wise to form the initial context:
\begin{equation}
\mathbf{c}_{\text{ctx}}
=
\mathrm{Concat}_{\text{ch}}
\Big(
\mathrm{Concat}_{t}\big[\mathbf{c}_{\text{env}},\,{\mathbf{c}_{\text{id}},\mathbf{c}_{\text{mem}}}\big],\,\mathbf{c}_{\text{mask}}
\Big).
\label{eq:ctx_seq}
\end{equation}
In the next subsection, we introduce a decoupled motion cross-attention module to inject motion conditions into $\mathbf{c}_{\text{ctx}}$, enabling compositional control over both pose dynamics and the global spatial placement of the human subject.

\begin{figure}[!t]
\centering
\includegraphics[width=0.95\linewidth]{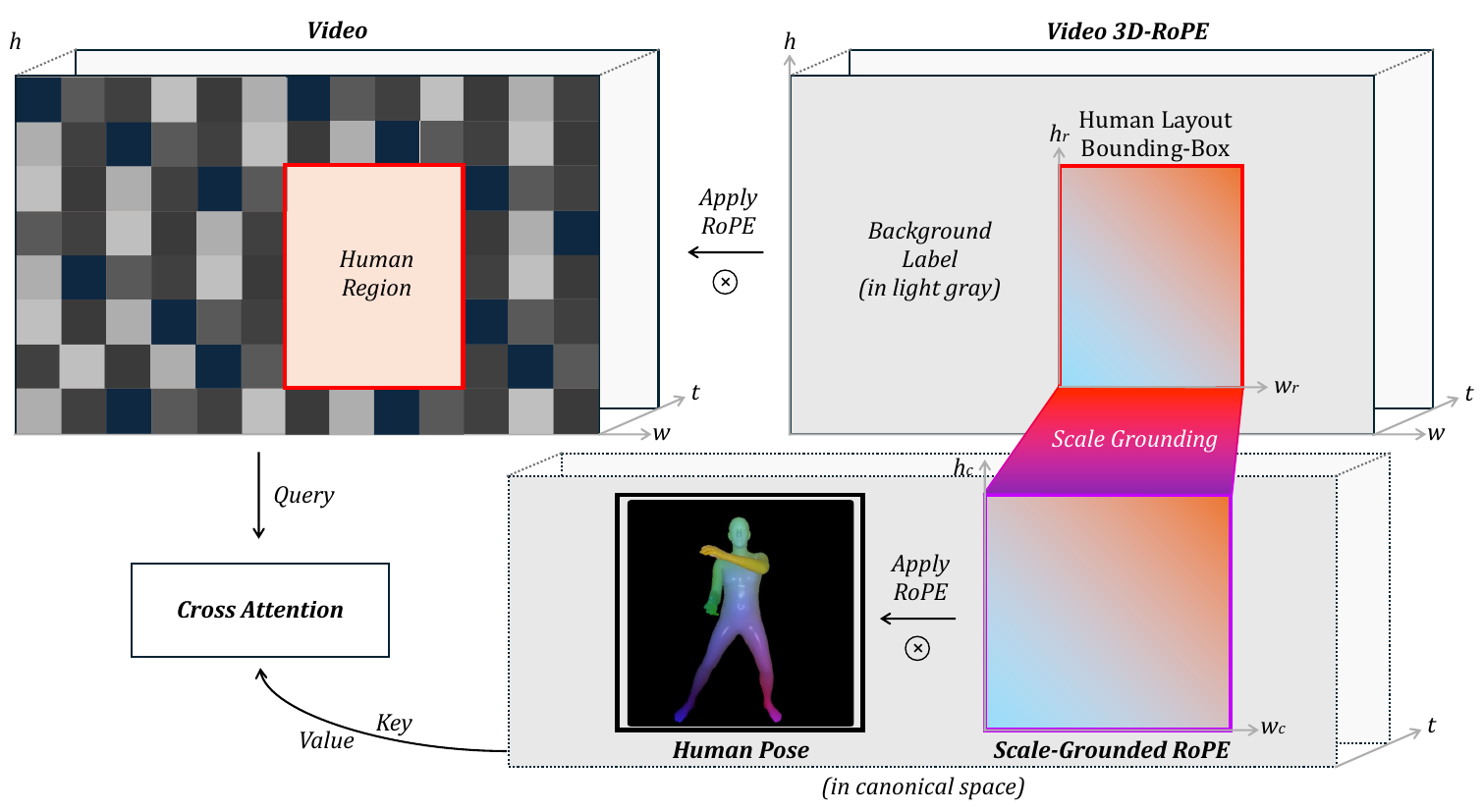}
\caption{\textbf{Decoupled Motion Cross-attention.}
The canonical-space human pose is injected into the video through cross-attention, where the proposed Dynamic-Grounded-RoPE bridges the spatial discrepancy between the environment and human spaces.
}

\label{fig:cross}
\end{figure}

\subsection{Decoupled Motion Cross-Attention}
\label{sec:motion_xattn}

To control both human motion and spatial placement, we decouple human control into 
(i) local motion represented in a canonical space and 
(ii) global placement specified by a coarse layout signal on the video token grid (e.g., bounding boxes). 
Fig.~\ref{fig:cross} illustrates the proposed cross-attention pathway for motion injection.

\noindent \textbf{Canonical-Space Human Motion Representation.}
We parameterize human motion using SMPL-X sequence $\mathbf{m}_{1:T}$.
To remove global displacement, we set the global translation to zero and center the body at a canonical origin.
We render the canonical SMPL-X mesh into a pose-aligned token map $\mathbf{u}_t\in\mathbb{R}^{h_c\times w_c\times C_u}$, where the rendering camera is chosen such that the body occupies a fixed proportion of the image (e.g., 90\%) to normalize scale across videos.
Stacking across time yields the motion tokens $\mathbf{c}_{\text{mot}} = \{\mathbf{u}_t\}_{t=1}^{T}$

\noindent \textbf{Dynamic-Grounded-RoPE.}
While the motion tokens $\mathbf{c}_{\text{mot}}=\{\mathbf{u}_t\}_{t=1}^{T}$ are formulated in a canonical space, 
the query features are defined on the video token grid, resulting in a scale mismatch between the two domains.
As shown in the red box of Fig.~\ref{fig:cross}, we introduce a per-frame placement region $\mathcal{B}_t$ on the video token grid to roughly indicate where the human should appear (see Sec.~\ref{sec:supp_2dbbox} of the Supplementary Material for details).
Its parameterization is given by $\boldsymbol{\beta}_t=(x^{1}_t,y^{1}_t,x^{2}_t,y^{2}_t)$, which defines the rectangular region $\mathcal{B}_t \;=\; \{(x,y)\mid x^{1}_t \le x < x^{2}_t,\; y^{1}_t \le y < y^{2}_t\}$.
Let $h_r = y^2_t - y^1_t$ and $w_r = x^2_t - x^1_t$ denote the height/width of $\mathcal{B}_t$ in token units, and recall $(h_c, w_c)$ as the spatial size of the canonical motion map.
We define per-frame scale factors that map video-grid coordinates to the canonical coordinates:
\begin{equation}
s^h_t = \frac{h_c}{h_r}, \qquad s^w_t = \frac{w_c}{w_r}.
\label{eq:scale_factor}
\end{equation}

For a query token at position $\mathbf{p}=(t,x,y)$ on the video grid, we apply a scale-grounded RoPE:
\begin{equation}
\tilde{\mathbf{Q}}(\mathbf{p}) =
\begin{cases}
\mathcal{R}\!\big(t,\, s^w_t x,\, s^h_t y\big)\,\mathbf{Q}(\mathbf{p}), & {(x, y)}\in\mathcal{B}_t,\\
\mathcal{R}\!\big(t,\, \alpha,\, \alpha\big)\,\mathbf{Q}(\mathbf{p}), & {(x, y)}\notin\mathcal{B}_t,
\end{cases}
\label{eq:q_rope}
\end{equation}
where $\alpha$ is a large constant (denoted as background label in Fig.~\ref{fig:cross}) used to collapse background positions into a shared embedding for tokens outside the target region.

For motion keys defined on the canonical grid, we apply the standard 3D RoPE without rescaling:
\begin{equation}
\tilde{\mathbf{K}}(\mathbf{p}_c)=\mathcal{R}\!\big(t,\, x_c,\, y_c\big)\,\mathbf{K}(\mathbf{p}_c),
\label{eq:k_rope}
\end{equation}
where $\mathbf{p}_c=(t,x_c,y_c)$ indexes positions on the canonical motion map.
Together, Eqs.~\eqref{eq:q_rope}--\eqref{eq:k_rope} calibrate positional encoding across the two decoupled domains, so that the target region on the video grid matches the canonical motion coordinates.

\noindent \textbf{Motion Injection by Cross-Attention.}
We inject motion via a dedicated cross-attention layer, where queries are taken from the video tokens and keys/values come from the canonical motion tokens.
Concretely, at frame $t$, we extract query features $\mathbf{Q}_t$ from the environmental features conditioned on $\mathbf{c}_{\text{env}}$.
We form keys and values from the canonical motion map $\mathbf{u}_t$ as $\mathbf{K}_t,\mathbf{V}_t$.
Applying the RoPE transformations in Eqs.~\eqref{eq:q_rope}--\eqref{eq:k_rope}, the motion injection is computed as
\begin{equation}
\mathrm{Attn}(\tilde{\mathbf{Q}}_t,\tilde{\mathbf{K}}_t,\mathbf{V}_t)
=
\mathrm{softmax}\!\left(\frac{\tilde{\mathbf{Q}}_t\tilde{\mathbf{K}}_t^{\top}}{\sqrt{d}}\right)\mathbf{V}_t,
\label{eq:motion_xattn}
\end{equation}
where $\tilde{\mathbf{Q}}_t$ is obtained by applying Eq.~\eqref{eq:q_rope} to the queries, and $\tilde{\mathbf{K}}_t$ is obtained by applying Eq.~\eqref{eq:k_rope} to the canonical motion keys.

\subsection{Model Training}
\label{sec:training}


\noindent \textbf{Multi-Task Training.}
A key advantage of our decoupled design is that environmental and motion supervision can be learned from heterogeneous sources.
We therefore adopt a multi-task masking strategy to train different components with partially available conditions.
Specifically, we use a binary mask sequence $\mathbf{c}_{\text{mask}}$ to indicate which region should be synthesized versus preserved, and construct the masked target in latent space as
\begin{equation}
\mathbf{z}^{\text{tgt}}_0
=
\mathbf{c}_{\text{mask}}\odot \mathbf{z}_0
+
(1-\mathbf{c}_{\text{mask}})\odot \mathbf{z}^{\text{keep}}_0,
\label{eq:mask_target}
\end{equation}
where $\mathbf{z}_0$ is the ground-truth video latent, $\mathbf{z}^{\text{keep}}_0$ denotes preserved latents, and $\odot$ is element-wise multiplication with broadcasting.

We consider three training modes:
(i) \textit{scene-only}, where we train the conditioning branch to reconstruct the environment using $\mathbf{c}_{\text{env}}$ (and optionally the dynamic context memory in $\mathbf{c}_{\text{mem}}$), while disabling motion injection;
(ii) \textit{motion-only}, where we train the motion cross-attention to synthesize the human inside the target region $\mathcal{B}_t$ using $\mathbf{c}_{\text{mot}}$, while preserving the outside region with $(1-\mathbf{c}_{\text{mask}})$;
and (iii) \textit{full conditioning}, where $\mathbf{c}_{\text{env}}, \mathbf{c}_{\text{mot}}, \mathbf{c}_{\text{id}}, \mathbf{c}_{\text{mem}}$ are jointly enabled.
This schedule encourages specialization of decoupled modules while maintaining compositional generalization at test time.
Empirically, we find that competitive controllability can be achieved with substantially less paired task data than fully-supervised fine-tuning. More training details are provided in Sec.~\ref{sec:supp_impl} of the Supplementary Material.

\noindent \textbf{Training objective.}
We follow flow-matching objective (Eq.~\eqref{eq:fm_loss_z}) and replace the data latent $\mathbf{z}_0$ with the masked target $\mathbf{z}^{\text{tgt}}_0$ in Eq.~\eqref{eq:mask_target}.

\section{Experiments}
\label{sec:experiments}

\subsection{Experimental Setup}
\label{sec:exp_setup}

\noindent \textbf{Datasets.}
As mentioned earlier, our method supports training with heterogeneous data sources.
We use a mixture of datasets that provide complementary supervision: (i) \textbf{EMDB2}~\cite{kaufmann2023emdbelectromagneticdatabaseglobal}, which contains sequences captured with dynamic cameras and provides paired global human/camera motion annotations for supervising human in dynamic scenarios; (ii) a human-motion subset of \textbf{MotionX}~\cite{lin2023motion}, which mainly consists of static-camera videos with diverse body motions; (iii) \textbf{ARKitScenes}~\cite{baruch2021arkitscenes}, which provides point-cloud captures used to strengthen environmental reconstruction and camera-aware projections; and (iv) \textbf{a self-collected set of web videos} to increase data diversity and long-horizon evaluation.
In total, our training set contains approximately {50k clips}, each {5--20 seconds} long.
For testing, we use \textbf{Traj100} following the same protocol as RealisMotion~\cite{liang2025realismotion}, as well as an additional self-collected cross-composition test set for evaluating swaps of identity, motion, and/or scene. More dataset details are provided in Sec.~\ref{sec:supp_data} of the Supplementary Material.


\noindent \textbf{Comparison methods.}
We compare with representative controllable human-video generation methods, including \textbf{RealisDance}, \textbf{Uni3C}, \textbf{WanAnimate}, and \textbf{RealisMotion}.
For each baseline, we follow the official inference settings and use the same prompts and evaluation protocol whenever applicable.

\noindent \textbf{Evaluation metrics.}
We evaluate both \textbf{self-reconstruction} (with paired ground truth) and \textbf{cross-composition} (without paired ground truth) using \textbf{FID} and \textbf{FVD} to measure visual quality and temporal realism, together with task-specific metrics including \textbf{Motion Smoothness} (MS), \textbf{Background Consistency} (BC), and \textbf{Subject Similarity} (SubSim).
For cross-composition, we additionally report \textbf{Face Similarity} (FaceSim) to assess identity swapping.
All metrics are computed using the same number of frames and the same resolution across methods.

\begin{table*}[t!] 
    \centering
    \caption{
    \textbf{Quantitative Results on Traj100 (Self-Reconstruction) and Our Constructed Testset (Cross-Composition).} 
    Lower is better for FID/FVD. Higher is better for others.
    }
    \label{tab:combined_results}
    \resizebox{1.0\linewidth}{!}{
    \begin{tabular}{l|ccccc|cccccc}
        \toprule
        & \multicolumn{5}{c|}{\textbf{Self-reconstruction (Traj100)}} & \multicolumn{6}{c}{\textbf{Cross-composition (Identity/Motion Swap)}} \\
        Method & FID$\downarrow$ & FVD$\downarrow$ & MS$\uparrow$ & BC$\uparrow$ & SubSim$\uparrow$ & FID$\downarrow$ & FVD$\downarrow$ & MS$\uparrow$ & BC$\uparrow$ & SubSim$\uparrow$ & Face$\uparrow$ \\
        \midrule
        WanAnimate~\cite{Wan-Animate}   & 21.25 & 224.08 & 0.893 & 0.956 & 0.189 & 152.07 & 1435.3 & 0.892 & 0.950 & \textbf{0.368} & 0.196 \\
        RealisDance~\cite{RealisDance-DiT}  & 22.17 & 247.54 & 0.920 & 0.946 & 0.237 & 155.71 & 1416.5 & 0.986 & 0.962 & 0.230 & \textbf{0.257} \\
        Uni3C~\cite{cao2025uni3c}        & 55.47 & 775.38 & 0.853 & 0.928 & 0.185 & 162.53 & 1519.5 & 0.924 & 0.950 & 0.321 & 0.203 \\
        RealisMotion~\cite{liang2025realismotion} & 18.89 & 193.89 & 0.916 & 0.940 & \textbf{0.356} & 151.19 & 1486.8 & 0.960 & 0.919 & 0.300 & 0.238 \\
        \midrule
        \rowcolor{lightgray!30} ONE-SHOT \textbf{(ours)} & \textbf{16.88} & \textbf{181.17} & \textbf{0.940} & \textbf{0.959} & {0.317} & \textbf{150.98} & \textbf{1382.4} & \textbf{0.993} & \textbf{0.963} & 0.349 & 0.241 \\
        \bottomrule
    \end{tabular}}
\end{table*}

\begin{figure*}[!t]
\centering
\includegraphics[width=0.95\linewidth]{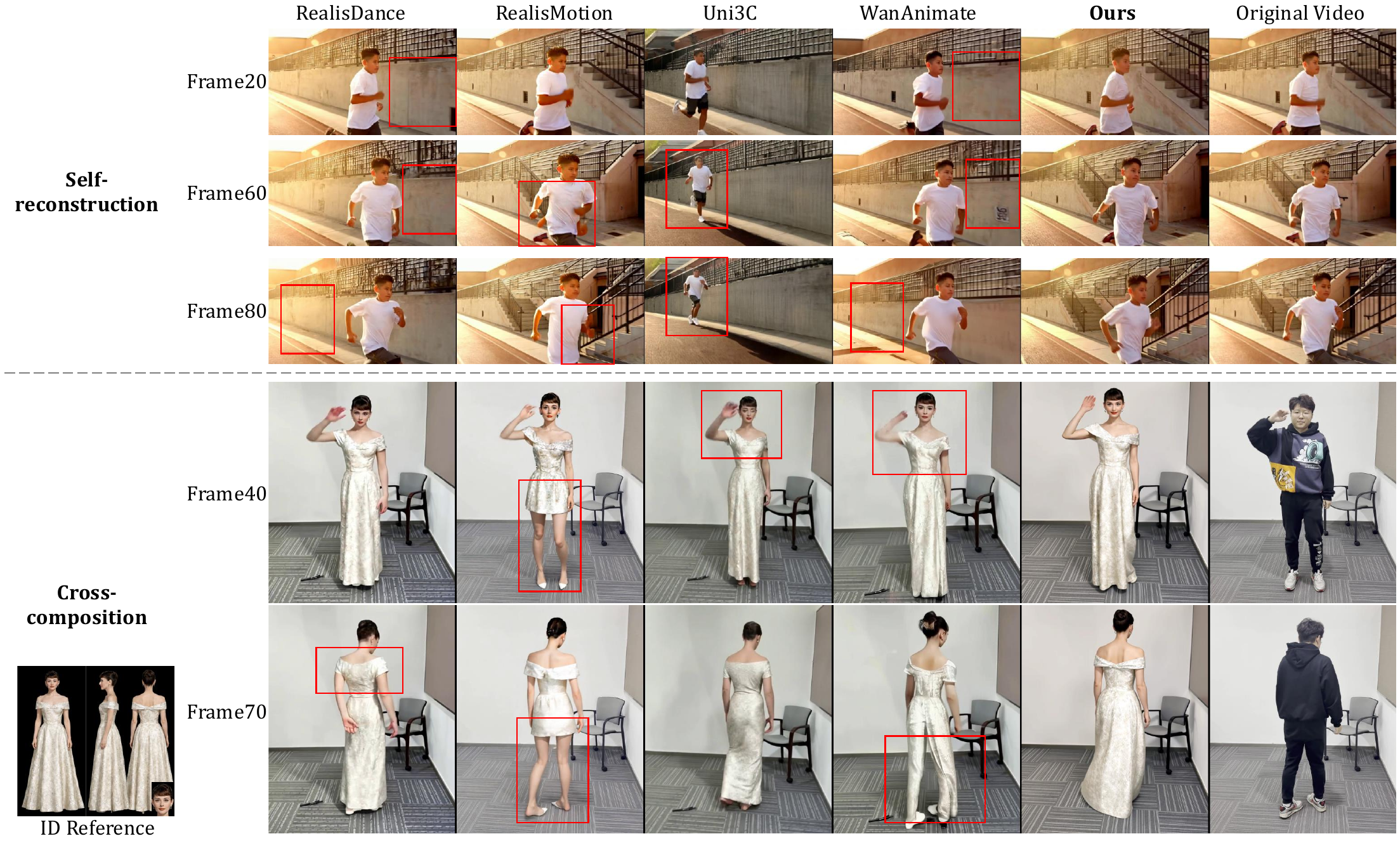}
\caption{
\textbf{Self-Reconstruction and Cross-Composition Comparisons on Traj100.} (a) Our method better preserves both the scene structure and human appearance, and (b) maintains stable subject placement and motion consistency when swapping identity.
}
\label{fig:self_cross}
\end{figure*}

\subsection{Evaluation Results}
\label{sec:exp_results}

\noindent \textbf{Quantitative results.}
Tab.~\ref{tab:combined_results} (left) reports self-reconstruction results on Traj100.
Our method achieves the best \textbf{FID} and \textbf{FVD}, indicating improved visual fidelity and temporal realism.
It also attains the highest \textbf{Motion Smoothness} and \textbf{Background Consistency}, suggesting that our decoupled design stabilizes both human motion synthesis and scene coherence.
While RealisMotion obtains a higher SubjectSim, our approach provides a more balanced trade-off across fidelity, motion stability, and background coherence, which is crucial for controllable human-environment generation.

Tab.~\ref{tab:combined_results}(right) evaluates cross-composition, where identity and motion are swapped across different scenes.
Our method achieves the best performance on \textbf{FID/FVD}, \textbf{Motion Smoothness} and \textbf{Background Consistency}, and remains competitive in \textbf{SubjectSim} and \textbf{FaceSim}, suggesting that our decoupled controls generalize beyond reconstruction.

\noindent \textbf{Qualitative results.}
Fig.~\ref{fig:self_cross} visualizes our results with prior methods under both self-reconstruction and cross-composition settings.
Our method produces more coherent backgrounds and more faithful motion dynamics, while baselines like WanAnimate and RealisDance primarily rely on the first-frame reference and do not explicitly model a persistent 3D-consistent environment, leading to background drifting, structural inconsistencies, or local collapse under camera motion. The results of RealisMotion contain blurred human motion and unrealistic swapped subjects accompanied by severe artifacts. Uni3C struggles to disentangle camera motion from human motion in most cases, resulting in weakened motion following where the subject appears nearly static or moves insufficiently.
In cross settings, we observe common failure modes such as {identity drift, background inconsistency, or motion mis-following}, whereas our decoupled motion injection maintains stable placement and consistent motion following.


\begin{figure}[!t]
\centering
\includegraphics[width=1.0\linewidth]{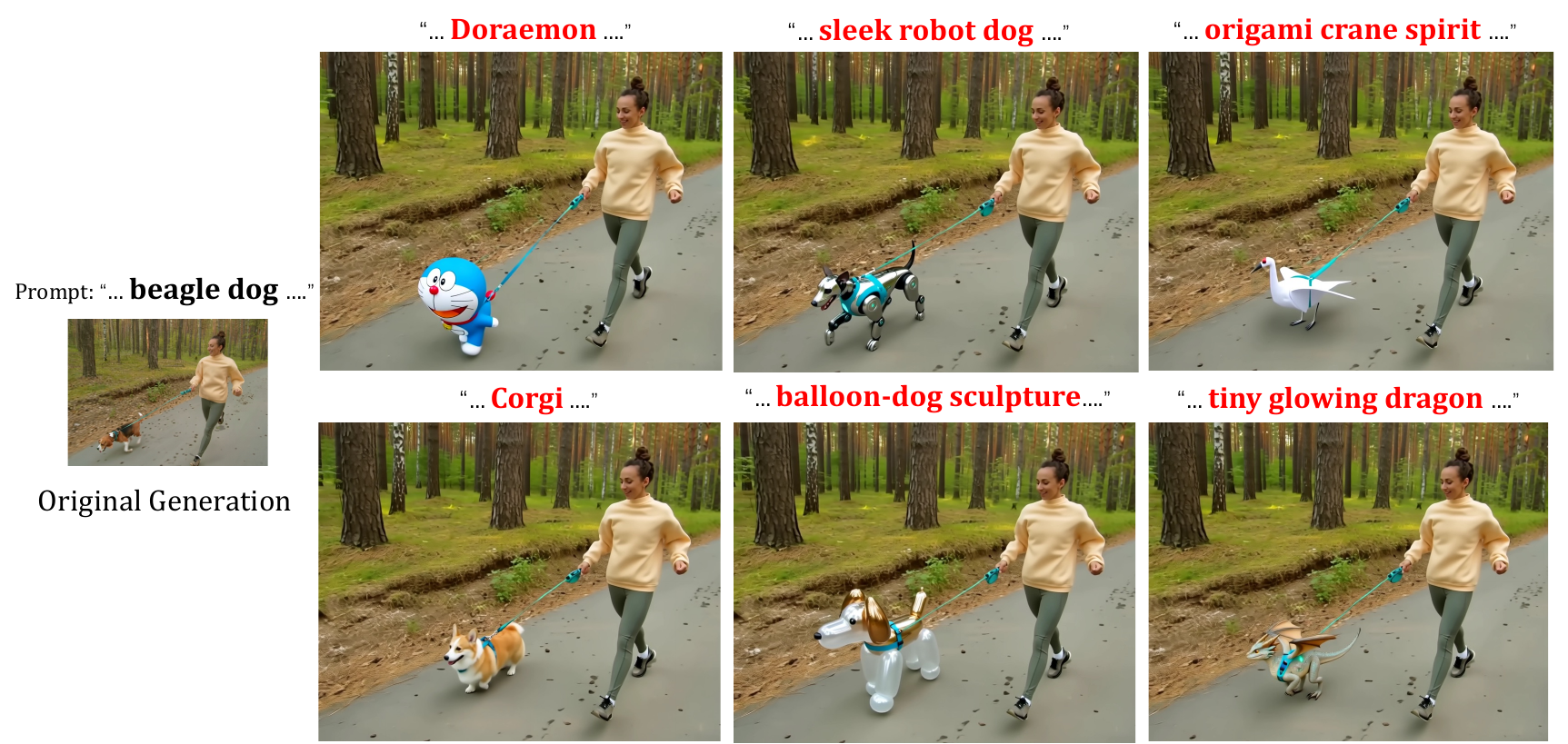}
\caption{
\textbf{Text-guided editing with compositional controls}.
Beyond composing scene, identity, and motion, our method enables instruction-based editing via text prompts, indicating strong compatibility with the pretrained VFM and minimal loss of its native text-conditioned editing ability.
}
\label{fig:prompt_editing}
\end{figure}

We can independently swap identity, motion, environment, and camera trajectory as shown in demo video, while also supporting instruction-based edits via text prompts as shown in Fig.~\ref{fig:prompt_editing}.
We also demonstrate the long video generation capability of our methods in the demo video and supplementary materials, where the scene and human appearance remain consistent over long horizons.

\subsection{Ablation Studies}
\label{sec:ablation}

\noindent \textbf{Dynamic-Grounded-RoPE.}
In Tab.~\ref{tab:ablation}, we ablate Dynamic-Grounded-RoPE by (i) removing scale grounding (2nd row) and (ii) replacing it with a naive label RoPE without 2D spatial coordinate--aware scaling (3rd row).
Both variants substantially degrade \textbf{FID/FVD} and reduce motion smoothness.
Without RoPE grounding, the model exhibits incorrect spatial placement and weaker motion following (see Fig.~\ref{fig:ab_rope}, 1st on the right), while naive label RoPE partially alleviates the issue but remains inferior to the full design.
These results indicate that 2D spatial coordinate--aware scaling is important for establishing stable spatial correspondence between canonical motion tokens and the target region on the environmental grid.

\begin{table}[t]
\centering
\caption{\textbf{Ablations on Traj100 (Self-Reconstruction).}}
\label{tab:ablation}
\begin{tabular}{lccc}
\toprule
Ablation & FID$\downarrow$ & FVD$\downarrow$ & MS$\uparrow$ \\
\midrule
\rowcolor{lightgray!30} ONE-SHOT \textbf{(ours)} & \textbf{16.88} & \textbf{181.17} & \textbf{0.9404} \\
\midrule
w/o RoPE grounding & 22.34 & 229.76 & 0.9268 \\
label-only RoPE & 21.36 & 221.66 & 0.9256 \\
w/o face reference & 21.37 & 215.43 & 0.9378 \\
w/o subject-cropped reference & 19.43 & 200.23 & 0.9351 \\
w/o {context memory} & 18.17 & 182.98 & 0.9330 \\
\bottomrule
\end{tabular}
\end{table}

\begin{figure}[t!]
\centering
\includegraphics[width=1.0\linewidth]{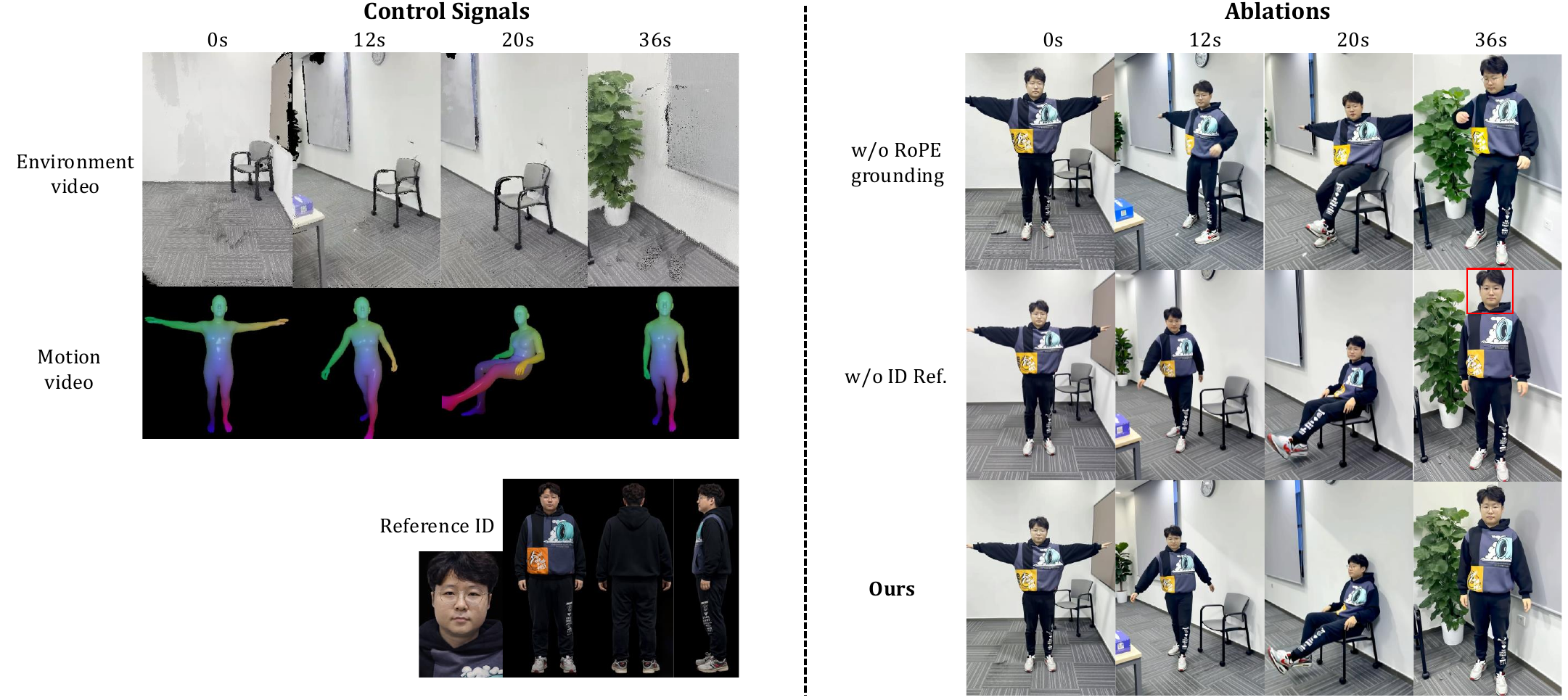}
\caption{
\textbf{Ablations of \textbf{Dynamic-Grounded-RoPE} and \textbf{Face Appearance Integration} on Long-Horizon Generation.}
Removing RoPE grounding weakens motion controllability, leading to pose drift and reduced adherence to the provided motion control.
Removing face reference weakens identity anchoring, causing increased identity drift over time.
Our full model achieves both better motion following and identity preservation.
}
\label{fig:ab_rope}
\end{figure}

\begin{figure}[!t]
\centering
\includegraphics[width=0.95\linewidth]{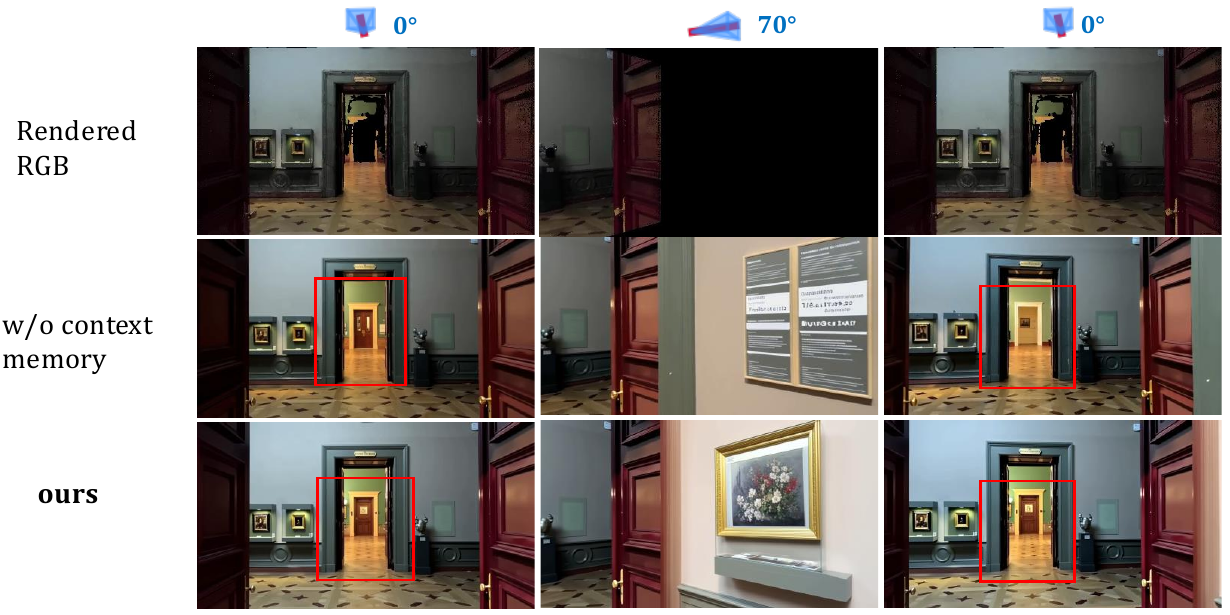}
\caption{
\textbf{Effect of context memory on viewpoint revisit consistency}.
%
Our model maintains structural and temporal consistency even after camera loops, whereas memoryless baselines suffer from appearance drift.
}
\label{fig:ab_memory}
\end{figure}

\noindent \textbf{Hybrid Context Integration.}
As shown in Tab.~\ref{tab:ablation}~(3rd-4th row) and Fig.~\ref{fig:ab_rope}~(2nd row on the right), removing face references or using uncropped references hurts reconstruction quality, suggesting that clean appearance cues are necessary for stable identity and texture rendering. 
Disabling context memory mainly affects long-horizon consistency: results become less stable and inconsistent when the camera revisits previous viewpoints like the case in Fig.~\ref{fig:ab_memory}.




\section{Limitation}
Our approach relies on the quality of the reconstructed scene point cloud and camera trajectory: inaccurate, sparse, or noisy point clouds (e.g., due to imperfect depth estimation or human removal) can reduce scene consistency and degrade generation quality. In addition, errors from the human estimation module (Human3R in this work) may also propagate to the final results. The method can also fail in extreme bbox-grounding cases (e.g., severely misplaced user-specified boxes), and some temporal drift may still accumulate in very long videos. Improving robust scene/human reconstruction, handling extreme bbox grounding, and fully eliminating long-horizon drift are important future directions, but are beyond the scope of this paper.
\section{Conclusion}

We introduce \textbf{ONE-SHOT}, a parameter-efficient framework for {compositional} human-environment video synthesis that targets three practical bottlenecks in prior human-centric VFMs: burdensome 3D alignment, over-conditioning that compromises generative flexibility, and short-horizon synthesis.
Our goal is to {factorize} human motion, environment geometry, camera trajectories, and human appearance into disentangled signals, and integrate them with minimal interference.
To this end, we proposed a \textbf{canonical-space motion injection} scheme that decouples human motion priors from environmental cues via cross-attention, explicitly reducing conditioning competition and preserving the VFM's instruction-following flexibility.
To eliminate the requirement of 3D alignment across disparate spatial domains, \textbf{Dynamic-Grounded-RoPE} establishes spatial correspondences directly in attention, avoiding fragile pre-alignment while retaining precise structural grounding.
Finally, \textbf{Hybrid Context Integration} (static reference tokens and dynamic memory tokens) sustains subject identity and scene coherence for \textbf{minute-level} long-horizon generation.
Extensive experiments and ablations show that ONE-SHOT achieves a more favorable balance between \textbf{structural control} and \textbf{creative diversity} than prior methods, while requiring only lightweight LoRA fine-tuning.

\clearpage

%
%
\bibliographystyle{splncs04}
\bibliography{main}

\clearpage
\markboth{}{}

\vspace*{0.5em}
\noindent{\LARGE\bfseries Supplementary Material\par}
\vspace{1em}

\noindent This supplementary material provides additional implementation details, data preprocessing descriptions, and extended ablations/qualitative results to support the main paper.

\setcounter{section}{0}
\renewcommand{\thesection}{\Alph{section}}
\renewcommand{\thesubsection}{\thesection.\arabic{subsection}}

\renewcommand{\theHsection}{supp.\Alph{section}}
\renewcommand{\theHsubsection}{supp.\Alph{section}.\arabic{subsection}}


\section{Implementation Details}
\label{sec:supp_impl}

\noindent \textbf{Backbone and trainable modules.}
Our model is built upon a pretrained video editing framework~\cite{vace} with a frozen VFM backbone, Wan2.1-14B~\cite{wanvideo}. 
We keep the backbone parameters fixed and update only lightweight trainable components: 
(i) standard LoRA adapters inserted into the conditioning branch, and 
(ii) the proposed motion cross-attention modules for controllable motion injection (Sec.~4.3 of the main paper). 
For LoRA, we follow a standard configuration and apply it to the query/key/value projections, output projection, and feed-forward layers (implemented as \texttt{to\_q}, \texttt{to\_k}, \texttt{to\_v}, \texttt{to\_out.0}, \texttt{ffn.net.0.proj}, and \texttt{ffn.net.2}). 
For motion injection, we additionally train the newly introduced motion-conditioning modules, including the extra motion cross-attention layer, the corresponding normalization layers, and the motion token embedding layer. 
Concretely, after the standard self-attention and text/image cross-attention, we append an additional cross-attention layer that takes motion tokens derived from SMPL-X as conditioning, together with our Dynamic-Grounded-RoPE for spatial grounding.

\noindent \textbf{Optimization and training setup.}
Training is performed at \textbf{$640\times640$}. 
We use different learning rates for the two trainable parts: \textbf{$1\times10^{-4}$} for LoRA parameters and \textbf{$1\times10^{-5}$} for the newly added motion cross-attention modules. 
To balance robustness and controllability, we adopt a mixed-conditioning schedule: \textbf{10\%} scene-only samples, \textbf{25\%} motion-only samples, and the remaining samples use full conditioning. 
To improve long-horizon generation, we additionally include motion-video context with \textbf{50\%} probability by using the first 1--9 frames as motion history, which stabilizes early temporal dynamics and reduces drift in long rollouts. We also apply extensive bbox augmentations (e.g., scaling and jittering), which encourage the model to treat the 2D bbox as a ``soft guide'' for human placement rather than a rigid constraint, thereby improving robustness to bbox misalignment during inference.



\section{Training Data and Pre-processing}
\label{sec:supp_data}



\noindent \textbf{Data sources.}
Our training videos are collected from both open-source datasets and self-collected human-centric footage, with an overall proportion of approximately \textbf{1:1}. 
The self-collected portion is designed to complement open-source data with diverse real-world compositions that are important for our task, especially outdoor human-motion videos with varied camera behavior and global motion patterns. 
To improve coverage, we deliberately balance several factors, including videos with and without camera motion, and videos where the subject remains roughly stationary versus exhibits clear global movement. 
This design helps expose ONE-SHOT to a broader range of compositional settings while remaining easy to reproduce in practice.

\noindent \textbf{Pre-processing philosophy.}
Our 3D pre-processing is intentionally lightweight and modular. 
Rather than requiring highly accurate scene reconstruction or explicit human-scene global alignment, we rely on standard off-the-shelf tools to obtain approximate decoupled scene and human signals. 
Moreover, because ONE-SHOT is trained in a decoupled and mixed-conditioning manner, not every training sample needs to provide every signal. 
For scene-only videos, we only extract scene-related conditions such as point clouds and camera trajectories; for motion-centric videos, especially some self-collected videos with static cameras, we primarily extract the human-motion signal. 
This flexibility reduces the dependence on fragile full-pipeline preprocessing.

\noindent \textbf{Decoupled signal construction.}
When full preprocessing is available, we reconstruct humans in world coordinates using Human3R~\cite{chen2025human3r} and estimate dense depth using Depth-Anything v3~\cite{lin2025depth3recoveringvisual}. 
We then recover camera extrinsics and build a background point cloud with the human removed. 
This process provides the conditioning inputs required by Sec.~4 of the main paper, including the background point cloud $\mathcal{P}$, camera trajectory $\{\mathbf{T}_t\}_{t=1}^{T}$, appearance references $\mathbf{y}^{\text{ref}}$ and optional memory $\mathbf{y}^{\text{mem}}$, and the target motion sequence $\mathbf{m}_{1:T}$.

\noindent \textbf{Test sets.}
For self-reconstruction, we use \textbf{Traj100} following the same protocol as RealisMotion~\cite{liang2025realismotion}. 
For cross-composition, we additionally construct a separate self-collected test set, since standard self-reconstruction benchmarks cannot evaluate generalization to novel identity-motion-scene recombination. 
This test set is built from 5 identities, 10 motions, and 10 scenes, with 83 plausible combinations selected for evaluation. 
It is used to evaluate compositional swaps of identity, motion, and/or scene.

\section{Details on 2D Bounding Box as Coarse Placement}
\label{sec:supp_2dbbox}

\noindent \textbf{Bbox during Training.}
During training, to ensure that ONE-SHOT treats the bounding box only as a soft spatial cue indicating where the sparse human should appear, we apply extensive bbox augmentations (e.g., scaling and jittering). This encourages the model to use the 2D prior as a flexible guide rather than a rigid constraint. As a result, ONE-SHOT remains robust to moderate bbox misalignment and can still synthesize natural foot-ground contact.

\noindent \textbf{Bbox during Testing.}
ONE-SHOT targets compositional generation by freely combining a new scene, human motion, and camera trajectory. Accordingly, we support flexible bbox configuration during inference, including automatically propagating a user-specified first-frame bbox, reusing detected bboxes from source footage when applicable, and even using fixed-size bboxes in tracking or hand-held shots:
\begin{enumerate}
    \item In general cases (e.g., most examples in our demo video), users specify a coarse square bbox in the first frame, denoted as $\mathcal{B}_0$, as a soft signal of the initial human placement. 
    The subsequent placement is then automatically determined by the source SMPL-X motion ${\{\boldsymbol{\Gamma}_t\}}_{t=1}^{T}$ and the target camera motion $\{\mathbf{T}_t\}_{t=1}^{T}$ and its intrinsic parameters.

    We define the bbox at frame $t$ as $\mathcal{B}_t = \{(u_t, v_t), a_t\}$, where $(u_t, v_t)$ is the bbox center and $a_t$ is the side length. Let the first-frame bbox be $\mathcal{B}_0 = \{(u_0, v_0), a_0\}$. Since the metric human height $H$ can be obtained from the SMPL-X motion ${\{\boldsymbol{\Gamma}_t\}}_{t=1}^{T}$, and in our canonical-space setting the human occupies a fixed proportion (e.g., $90\%$) of the bbox side length, the first-frame root depth can be easily estimated by perspective projection as
    \begin{equation}
        Z_0 = \frac{f \cdot H}{0.9 a_0},
    \end{equation}
    where $f$ is the focal length of the target camera.

    Given the first-frame bbox $\mathcal{B}_0$, we can then back-project the 2D bbox center to obtain the initial 3D human root location $P_0=(X_0,Y_0,Z_0)$:
    \begin{equation}
        P_0 = (X_0,Y_0,Z_0) = \left(\frac{(u_0-c_x)\cdot Z_0}{f}, \frac{(v_0-c_y)\cdot Z_0}{f}, Z_0\right),
    \end{equation}
    where $c_x, c_y,$ and $f$ are the intrinsic parameters of the target camera.

    After obtaining the first-frame 3D human root location, we propagate the root trajectory by following the relative motion of the desired SMPL-X sequence:
    \begin{equation}
        \{P_t\}_{t=1}^{T} = \{P_0 + \Delta \boldsymbol{\Gamma}_t\}_{t=1}^{T},
    \end{equation}
    where $\{\Delta \boldsymbol{\Gamma}_t\}_{t=1}^{T}$ denotes the relative root motion derived from the source SMPL-X motion $\{\boldsymbol{\Gamma}_t\}_{t=1}^{T}$.

    For each following frame, the propagated 3D root is projected under the target camera extrinsics $\{\mathbf{T}_t\}_{t=1}^{T}$ to obtain the bbox center, while the bbox scale is updated according to the projected depth:
    \begin{equation}
        \big\{\mathcal{B}_t\big\}_{t=1}^{T} \;=\; \Pi\!\left(\{P_t\}_{t=1}^{T}; \{\mathbf{T}_t\}_{t=1}^{T}\right),
    \end{equation}
    where $\Pi(\cdot)$ denotes the bbox construction procedure based on projection and depth-dependent scaling. This provides a simple and automatic way to obtain coarse bboxes for arbitrary human motion under arbitrary target camera trajectories, while the only user-specified factor is the coarse first-frame placement.

    \item ONE-SHOT is also flexible when users do not want to manually define a bbox. A simple workaround is to reuse detected bboxes when the target scene is derived from footage that originally contained a person. If the target camera trajectory remains the same as that source scene video (e.g., video reenactment), the following bboxes can also be directly reused. Otherwise, they can still be automatically calculated as above under a new target camera trajectory.

    \item In other scene-only cases, such as tracking or hand-held shots, the bbox can even remain static across frames because the camera motion itself already mimics a following perspective. This is significantly more user-friendly than previous methods requiring complex and fragile 3D pre-processing.
\end{enumerate}

\section{More Discussion}\label{sec:extradiscussion}
In this section, we provide additional discussion about our method as follows:

\noindent \textbf{1) Fair comparison with baselines.}
For all baselines, we use their official checkpoints and inference pipelines. Best-effort adaptation is only applied when their interface does not match our compositional task with disentangled control of motion, scene, camera, and identity. 
The trajectory shift observed in some baselines is not caused by a different target camera trajectory, but by their fragile human-scene global alignment, where scale or magnitude errors can lead to unstable placement or apparent camera mismatch.
To further strengthen fairness, we also perform a best-effort WAN-VACE adaptation by fine-tuning it on the same data used by ONE-SHOT so that it can take SMPL-X as compositional control input. ONE-SHOT remains clearly better (see Tab.~\ref{tab:vace_comparison}). WAN-VACE also shows weaker identity preservation and more frequent unstable human scale, placement mismatch, and weaker human-scene alignment.

\begin{table}[h!]
    \centering
    \caption{
    \textbf{Comparison with fine-tuned WAN-VACE on self-reconstruction and cross-composition.}
    WAN-VACE is adapted to our setting by fine-tuning on the same training data as ONE-SHOT, with SMPL-X used as compositional control input. Lower is better for FID/FVD. Higher is better for others.
    }
    \label{tab:vace_comparison}
    \resizebox{0.98\linewidth}{!}{
    \begin{tabular}{l|ccccc|cccccc}
        \toprule
        & \multicolumn{5}{c|}{\textbf{Self-reconstruction (Traj100)}} & \multicolumn{6}{c}{\textbf{Cross-composition}} \\
        Method & FID$\downarrow$ & FVD$\downarrow$ & MS$\uparrow$ & BC$\uparrow$ & SubSim$\uparrow$ & FID$\downarrow$ & FVD$\downarrow$ & MS$\uparrow$ & BC$\uparrow$ & SubSim$\uparrow$ & Face$\uparrow$ \\
        \midrule
        WAN-VACE (fine-tuned) & 20.96 & 185.09 & 0.935 & 0.955 & 0.311 & 151.68 & 1465.3 & 0.912 & 0.947 & 0.182 & 0.099 \\
        \rowcolor{lightgray!30} ONE-SHOT \textbf{(ours)} & \textbf{16.88} & \textbf{181.17} & \textbf{0.940} & \textbf{0.959} & \textbf{0.317} & \textbf{150.98} & \textbf{1382.4} & \textbf{0.993} & \textbf{0.963} & \textbf{0.349} & \textbf{0.241} \\
        \bottomrule
    \end{tabular}}
\end{table}

\noindent \textbf{2) How faithfully camera trajectories and environmental layout are preserved.}
ONE-SHOT uses per-frame RGB-D projections rendered from the target point cloud under the target camera trajectory, so camera motion and scene layout are preserved through explicit geometric conditioning.

\noindent \textbf{3) Dependence on 3D preprocessing / pipeline clarity.}
ONE-SHOT uses standard off-the-shelf preprocessing for scene and human reconstruction, as discussed in Sec.~\ref{sec:supp_data}, rather than fragile 3D global alignment as in prior methods. We further apply augmentation/dropout on scene control during training and support pure RGB scene input at inference, making the model less sensitive to scene control quality.

\section{Border Impact}
This study advances the field of controllable video generation by providing a parameter-efficient framework for compositional human-environmental video synthesis. While these generative capabilities empower creative expression and long-horizon production, they may inadvertently facilitate the creation of misleading information or fabricated visuals, necessitating high vigilance regarding potential misuse. Furthermore, as our model is built upon extensive datasets, it is crucial to address concerns of privacy and consent, as well as the risk of reinforcing underlying data biases that could lead to unjust outcomes. We advocate for the responsible and inclusive deployment of these technologies and emphasize that our work focuses on technical innovation using publicly available models and datasets.

\end{document}